%% file: paper.tex
\documentclass{article}
\usepackage{microtype}
\usepackage{graphicx}
\usepackage{subfig}
\usepackage{booktabs} 

\usepackage{hyperref}

\usepackage[accepted]{icml2023}

\usepackage{amsmath}
\usepackage{amssymb}
\usepackage{mathtools}
\usepackage{amsthm}
\usepackage{multirow}

\usepackage[capitalize,noabbrev]{cleveref}

\usepackage{paralist, tabularx}

\theoremstyle{plain}
\newtheorem{theorem}{Theorem}[section]

\theoremstyle{definition}
\newtheorem{definition}[theorem]{Definition}

\theoremstyle{remark}

\usepackage[textsize=tiny]{todonotes}

\icmltitlerunning{Multiresolution Graph Transformers and Wavelet Positional Encoding for Learning Hierarchical Structures}

\begin{document}

\twocolumn[
\icmltitle{Multiresolution Graph Transformers and Wavelet Positional Encoding for Learning Hierarchical Structures}



\icmlsetsymbol{equal}{*}

\begin{icmlauthorlist}
\icmlauthor{Nhat Khang Ngo}{equal,fpt}
\icmlauthor{Truong Son Hy}{equal,ucsd}
\icmlauthor{Risi Kondor}{uchicago}
\end{icmlauthorlist}

\icmlaffiliation{fpt}{FPT Software AI Center, Hanoi, Vietnam}
\icmlaffiliation{ucsd}{Halıcıoğlu Data Science Institute, University of California San Diego, La Jolla, USA}
\icmlaffiliation{uchicago}{Department of Computer Science, University of Chicago}

\icmlcorrespondingauthor{Truong Son Hy}{tshy@ucsd.edu}

\icmlkeywords{Machine Learning, ICML}

\vskip 0.3in
]



\printAffiliationsAndNotice{\icmlEqualContribution} 

\input{abstract}
\input{introduction}
\input{related}

\input{background}
\input{method}
\input{experiments}

\input{conclusion}
\input{acknowledgements}

\bibliography{paper}
\bibliographystyle{icml2023}

\newpage
\appendix
\onecolumn

\input{appendix}


\end{document}

%% file: abstract.tex
\begin{abstract}
Contemporary graph learning algorithms are not well-defined for large molecules since they do not consider the hierarchical interactions among the atoms, which are essential to determine the molecular properties of macromolecules. In this work, we propose Multiresolution Graph Transformers (MGT), the first graph transformer architecture that can learn to represent large molecules at multiple scales. MGT can learn to produce representations for the atoms and group them into meaningful functional groups or repeating units. We also introduce Wavelet Positional Encoding (WavePE), a new positional encoding method that can guarantee localization in both spectral and spatial domains. Our proposed model achieves competitive results on two macromolecule datasets consisting of polymers and peptides, and one drug-like molecule dataset. Importantly, our model outperforms other state-of-the-art methods and achieves chemical accuracy in estimating molecular properties (e.g., GAP, HOMO and LUMO) calculated by Density Functional Theory (DFT) in the polymers dataset. Furthermore, the visualizations, including clustering results on macromolecules and low-dimensional spaces of their representations, demonstrate the capability of our methodology in learning to represent long-range and hierarchical structures. Our PyTorch implementation is publicly available at \url{https://github.com/HySonLab/Multires-Graph-Transformer}.

\vspace{-15pt}
\end{abstract}

%% file: introduction.tex
\section{Introduction} \label{sec:intro}

Macromolecules are long-range and hierarchical structures as they consist of many substructures. While small molecules in existing datasets \cite{Ruddigkeit2012, Ramakrishnan2014, Sterling2015} comprise less than 50 atoms connected by simple rings and bonds, this number in a macromolecule can be dozens or even hundreds. Substructures such as repeating units and functional groups are intrinsic parts of macromolecules; they present unique chemical reactions regardless of other compositions in the same molecules \cite{functional}. Therefore, studying the multiresolution characteristic of large molecules is imperative to gain comprehensive knowledge about real-life materials like polymers or proteins \cite{multiscale}. In recent years, several works \cite{machine_learning_exploit, screening_macro, macro_molecule} have been proposed to apply machine learning algorithms to learn macromolecules at multiple scales. These approaches, however, rely on thorough feature selection and extraction, which are inefficient when learning from large databases of multicomponent materials \cite{machine_learning_exploit}.


Message passing is a prevailing paradigm for designing neural networks that operate on graph-structured data. Previous studies \cite{PMLR-V70-gilmer17a, gcn, gat, ppa, gin} have proposed different strategies to perform message passing on graphs and achieved remarkable results across various domains. However, message-passing-dominated graph neural networks (GNNs) have some inherent limitations, such as limited expressiveness capability \cite{high-order,gin}, over-smoothing \cite{chen-oversmoothing, li-2018, oono2020graph}, over-squashing  \cite{alon2021on} issues. Over-smoothing exists in graph neural networks that consist of a sufficiently large number of layers, and node representations are likely to converge to a constant after going through these deep networks. Over-squashing problems occur when messages are ineffectively propagated and aggregated through bottlenecks on long-range graph structures. These two shortcomings hinder GNNs from making good predictions on long-range and hierarchically structured data. Furthermore, the molecular properties of large molecules are formed not only by interactions among atoms within neighborhoods but also by distant atoms. Therefore, local information is not sufficient to model macromolecules.


Transformers are classes of deep learning models that leverage self-attention mechanisms to handle long-range dependencies in various data domains, such as natural language processing \cite{vaswari, bert} or computer vision \cite{vitrans, Swin}. In graph domains, Transformer-like architectures \cite{san, generalization_trans, gps} have proved their effectiveness in learning node representations as they can overcome the over-smoothing and over-squashing issues by directly measuring the pairwise relationships between the nodes. Contrary to GNNs, graph transformers do not use the graph structure as hard-coded information. They, instead, encode positional and structural information on graphs as soft inductive bias, making them flexible learners in graph learning problems \cite{san}. Node positional representations can be derived based on spectral \cite{graph_generalize, rw} or spatial \cite{anchor, distance} domains. Most existing spectral-based methods decompose the graph Laplacian into sets of eigenvectors and eigenvalues. However, these eigenvectors have sign ambiguity and are unstable due to eigenvalue multiplicities. On the other hand, spatial-based approaches compute the shortest distances among the nodes; however, these encoding methods do not consider the structural similarity between nodes and their neighborhoods \cite{sat}.

We propose Multiresolution Graph Transformer (MGT) and 
Wavelet Positional Encoding (WavePE), using multiresolution analysis on both spectral and spatial domains for learning to represent hierarchical structures. Our contributions are four-fold:
\begin{itemize}
    \item We design Multiresolution Graph Transformer (MGT), a Transformer-like architecture that can operate on macromolecules at multiple scales. Our proposed model can learn the atomic representations and group them into meaningful clusters via a data-driven algorithm. Finally, the substructures, i.e. clusters, are fed to a Transformer encoder to calculate the representations of several substructures in macromolecules.
    \item We introduce Wavelet Positional Encoding (WavePE), a new positional encoding scheme for graph-structured data. Since wavelet analysis can provide localization in both spatial and spectral domains, we construct a set of wavelets to capture the structural information on graphs at different scales. Then, we apply equivariant encoding methods to project the wavelet tensors into positional representations for the atoms.
    \item We show the effectiveness of our methodology by reporting its superior performance on three molecular property prediction benchmarks. These datasets contain macromolecules, i.e. peptides and polymers, that are highly hierarchical and consist of up to hundreds of atoms. Our model achieves important chemical accuracy in DFT approximation for the polymers dataset.
    \item Our visualization demonstrates the comprehensiveness of our proposed methods in learning to represent large molecules. In general, we show the representations of molecules produced by MGT and how MGT determines and groups the atoms in long-chain molecules.
\end{itemize}

%% file: related.tex
\section{Related work} \label{sec:related}
\paragraph{Hierachical Learning on Molecules}
Functional groups or repeating units are essential phenomena in chemistry. While functional groups constitute large molecules, repeating units are the primary parts that produce complete polymer chains. We regard them as substructures. In particular, similar substructures undergo similar chemical reactions regardless of the remaining compositions existing in the molecules \cite{functional}. Previous work has leveraged the hierarchical property of molecules to improve the performance in molecular representation learning and generation. \citet{knowledge_constrastive}, \citet{grover}, and \citet{hyper_message} used functional groups as prior knowledge to guide the models to predict accurate molecular properties. For the molecular generation task, \citet{structural_motifs} and \citet{hpgraph} used chemical rules to extract substructures and constructed a vocabulary of structural motifs to generate the molecules.

\paragraph{Graph Transformers} Earlier research efforts have adopted Transformer-like architectures to graph-structured data. \citet{graph_generalize} proposed an early approach to generalize Transformers to graphs using Laplacian positional encoding and performing self-attention on one-hop neighbors surrounding center nodes. On the other hand, \citet{san} computes attention scores on the entire graph with differentiation between positive and negative edges, while also using Laplacian positional encoding. \citet{grover} introduced GTransformer that utilizes vectorized outputs from local GNNs as inputs for a Transformer encoder, making up an effective  combination between node local and global information. \citet{gps} proposed a general framework that integrates essential components of Graph Transformers, including positional or structural encoding, graph feature extraction, local message passing, and self-attention. Also, \citet{sat} extracted multiple k-hop subgraphs and fed them to local GNNs to compute their embeddings, which are then moved to a Transformer encoder. Graphormer proposed in \cite{graphormer} uses attention mechanisms to estimate several types of encoding, such as centrality, spatial, and edge encodings. In addition, \citet{pure_transformer} treated all nodes and edges as independent tokens augmented with orthonormal node identifiers and trainable type identifiers, and fed them to a standard Transformer encoder. 
Moreover, \citet{graph_transformer_networks} generated multiple meta-paths, i.e. views, of a graph and computed their pairwise attention scores, before aggregating them into a final representation for the entire graph.

\paragraph{Graph Positional Encoding} Several approaches have been proposed to encode the positional or structural representations into node features to improve the expressiveness of GNNs and Graph Transformers. Node positions can be determined via spectral or spatial domains. Spectral-based methods include Laplacian positional encoding \cite{graph_generalize, san} and random walk positional encoding (RWPE) \cite{rw}. For spatial-based methods, \citet{anchor} computed distances of sets of nodes to anchor nodes, whereas \citet{distance} calculated the shortest distances between pairs of nodes.

\paragraph{Multiresolution Analysis and Wavelet Theory} 
Multiresolution Analysis (MRA) has been proposed by \cite{192463, 10.5555/1525499} as a method to approximate signals at multiple scales in which the signals are decomposed over elementary waveforms chosen from a family called wavelets (i.e. mother wavelets and father wavelets), including Haar \cite{Haar1910ZurTD}, Daubechies \cite{Daubechies1988OrthonormalBO}, etc., to produce the sparse representations. In graph and discrete domains, \citet{HAMMOND2011129} introduced spectral graph wavelets that are determined by applying the wavelet operator on the graph Laplacian at multi-levels. \citet{COIFMAN200653} proposed diffusion wavelet that is a fast multiresolution framework for analyzing functions on discretized structures such as graphs and manifolds. In the deep learning era, \citet{10.5555/2999611.2999723} and \citet{xu2018graph} leveraged the power of neural networks for graph wavelet construction and computation.



%% file: background.tex
\section{Background} \label{sec:background}

\subsection{Notation}
A molecule can be represented as an undirected graph in which nodes are the atoms and edges are the valency bonds between them. In paticular, we refer to a molecular graph as $G = (\mathcal{V}, \mathcal{E}, \mathbf{A}, \mathbf{X}, \mathbf{P}, \mathcal{V}_s)$, where $G$ is an undirected graph having $\mathcal{V}$ ($|\mathcal{V}| = n$) and $\mathcal{E}$ as sets of nodes and edges respectively; also, $\mathbf{A} \in \mathbb{R} ^ {n \times n}$ is the graph's adjacency matrix. When a graph is attributed, we augment $G$ with a set of node feature vectors $\mathcal{X} = \{x_1,..., x_n\}, x_i \in \mathbb{R}^d$ and a set of node positional vectors $\mathcal{P} = \{p_1, ..., p_n\}, p_i \in \mathbb{R}^p$. These two types of attributes are stored in $\mathbf{X} \in \mathbb{R} ^ {n \times d}$ and $\mathbf{P} \in \mathbb{R} ^ {n \times p}$ correspondingly.
In addition to the atom-level representation of $G$, $\mathcal{V}_s = \{v_{s_1}, ... , v_{s_k}\}$ denotes the substructure set in which $v_{s_i} \subset \mathcal{V}$, i.e. $v_{s_i}$ is a subset of atoms of the molecule.

\subsection{Hierachical Learning on Molecules}
\label{heirachical_coarsen}
Molecular property prediction is regarded as a graph-level learning task. We need to aggregate node embeddings into graph-level vectors which are then fed to a classifier to make predictions on graphs. Specifically, a function $f: \mathcal{V} \xrightarrow{} \mathcal{Z}$ that maps the atom $u \in \mathcal{V}$ to a $d_o$-dimensional vector $z_u \in \mathcal{Z} \subset \mathbb{R} ^ {d_o}$ should learn to produce atom-level representations. Most existing graph neural networks compute the vector $z =\zeta (\{f(u) | u \in \mathcal{V}\})$ that indicates a representation for the entire molecular graph, where $\zeta$ can be sum, mean, max, or more sophisticated operators. For hierarchical learning, substructure-level representations can be derived in addition to atom-level representations by aggregating node representations in the same substructures as $z_s =\zeta (\{f(u) | u \in v_s \land v_s \in \mathcal{V}_s\})$. Instead of atom vectors, we aggregate the substructure vectors to represent the entire graph, i.e. $z = \zeta (\{z_s | z_s \in \mathcal{V}_s\})$. Finally, a classifier $g$ given $z$ as inputs is trained to predict the molecular properties.


\subsection{Transformers on Graphs}
While GNNs learn node embeddings by leveraging the graph structure via local message-passing mechanisms, Transformers disregard localities and directly infer the relations between pairs of nodes using only node attributes. In other words, the node connectivity is not utilized in pure transformer-like architectures \cite{vaswari}, reducing the graph conditions to a set learning problem. Given a tensor of node features $\mathbf{X} \in \mathbb{R} ^ {n \times d}$, Transformers compute three matrices including query ($\mathbf{Q}$), key ($\mathbf{K}$), and value ($\mathbf{V}$) via three linear transformations  $\mathbf{Q} = \mathbf{X} \mathbf{W}_q^T$, $\mathbf{K} = \mathbf{X} \mathbf{W}_k ^ T$, and $\mathbf{V} = \mathbf{X} \mathbf{W}_v ^ T$. A self-attention tensor ($\mathbf{H}$) can be computed as follows:
\begin{equation}
    \mathbf{H} = \text{softmax}(\frac{\mathbf{Q}\mathbf{K}^T}{\sqrt{d_o}}) \mathbf{V}
    \label{eq:1}
\end{equation}
where $\mathbf{W}_q$, $\mathbf{W}_k$, and $\mathbf{W}_v$ are learnable parameters in $\mathbb{R} ^ {d_o \times d}$, resulting in $\mathbf{H} \in \mathbb{R} ^{n \times d_o}$. Furthermore, each $\mathbf{H}$ in Eq. \ref{eq:1} denotes an attention head. To improve effectiveness, multiple $\{\mathbf{H}\}_{i = 1}^ h$ are computed, which is known as multi-head attention. All of the attention heads are concatenated to form a final tensor: $\mathbf{H}_o = \text{concat}(\mathbf{H}_1,..., \mathbf{H}_h) $, where $h$ is the number of attention heads. Finally, the output $\mathbf{X}^\prime$, i.e. new node representations, can be computed by feeding $\mathbf{H}_o$ into a feed-forward neural network (FFN), i.e. $\mathbf{X} ^\prime = \text{FFN}(\mathbf{H}_o)$. It is easy to see that Transformers operating on inputs without positional encoding are permutation invariant.

\paragraph{Positional Encoding} As pure Transformer encoders only model sets of nodes without being cognizant of the graph structures, positional or structural information between nodes and their neighborhoods should be incorporated into node features. In particular, node positional representations can be added or concatenated with node features, resulting in comprehensive inputs for Transformer-like architectures operating on graph-structured data.

%% file: method.tex
\begin{figure*}[t]
\centering
\captionsetup[subfloat]{captionskip=5mm} 
\subfloat[]{%
\label{fig:wavepe}%
\includegraphics[scale = 0.4]{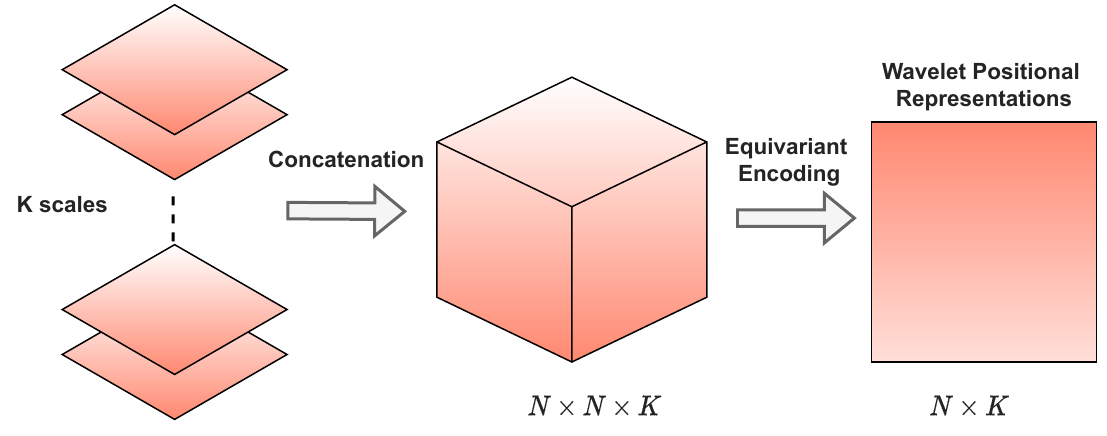}}%
\qquad
\subfloat[]{%
\label{fig:atom_level_mgt}%
\includegraphics[scale = 0.47]{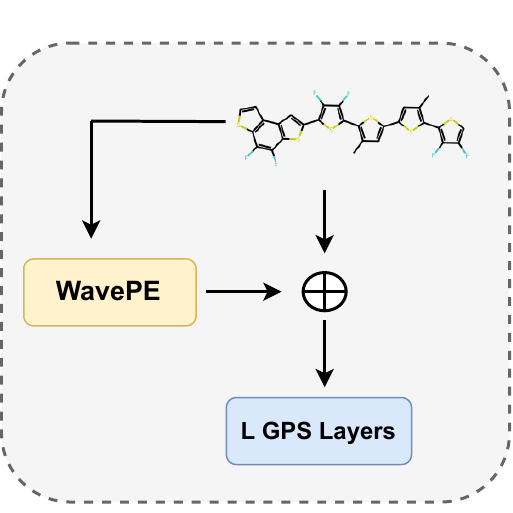}}%
\qquad
\subfloat[]{%
\label{fig:substructure_level_mgt}%
\includegraphics[scale = 0.32]{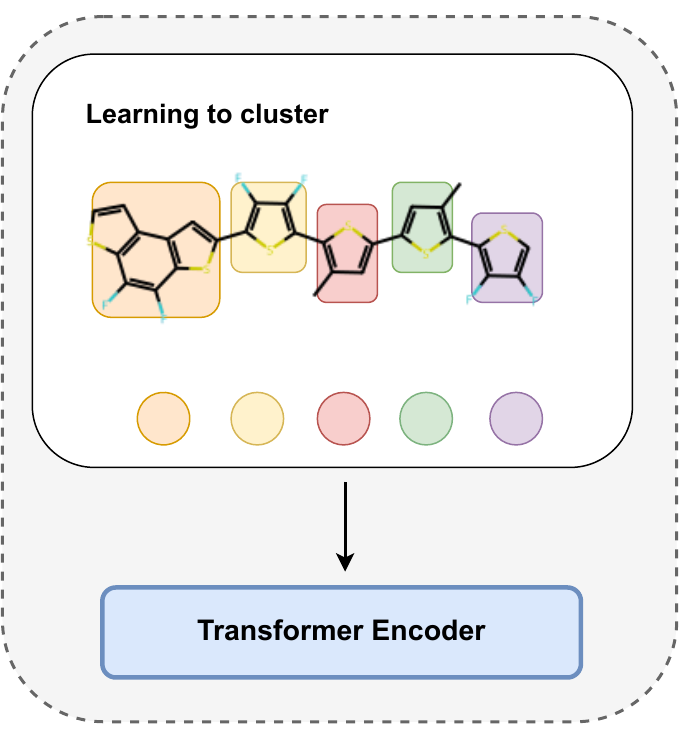}}%
\caption{Overview of Wavelet Positional Encoding (WavePE) and Multiresolution Graph Transformer (MGT). a) $k$ diffusion matrices of size $N \times N$ are stacked together to produce a wavelets tensor with size $N \times N \times K$ which are contracted by equivariant encoding methods to yield a tensor of positional representation $N \times k$. b) Atomic representations are derived by passing the molecular graph augmented with positional features through $L$ GPS layers. c) A macromolecule is decomposed into several substructures in which the features are aggregated from the atom-level outputs, resulting in a set of substructures that are moved to a Transformer encoder.}
\label{fig:overall-idea}
\end{figure*}

\section{Wavelet Positional Encoding}
\label{wavelet}

\subsection{Spectral Graph Wavelets}
\begin{figure}%
\centering
\captionsetup[subfloat]{captionskip=-2mm} 
\subfloat[s = 2]{%
\label{fig:peptides}%
\includegraphics[scale = 0.2]{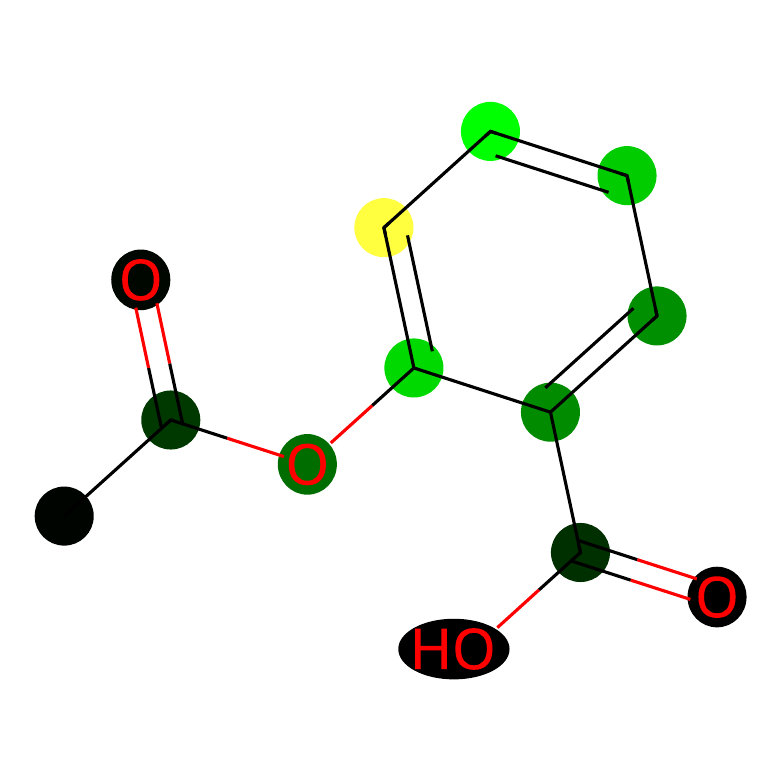}}%
\subfloat[s = 3]{%
\label{fig:polymer}%
\includegraphics[scale = 0.2]{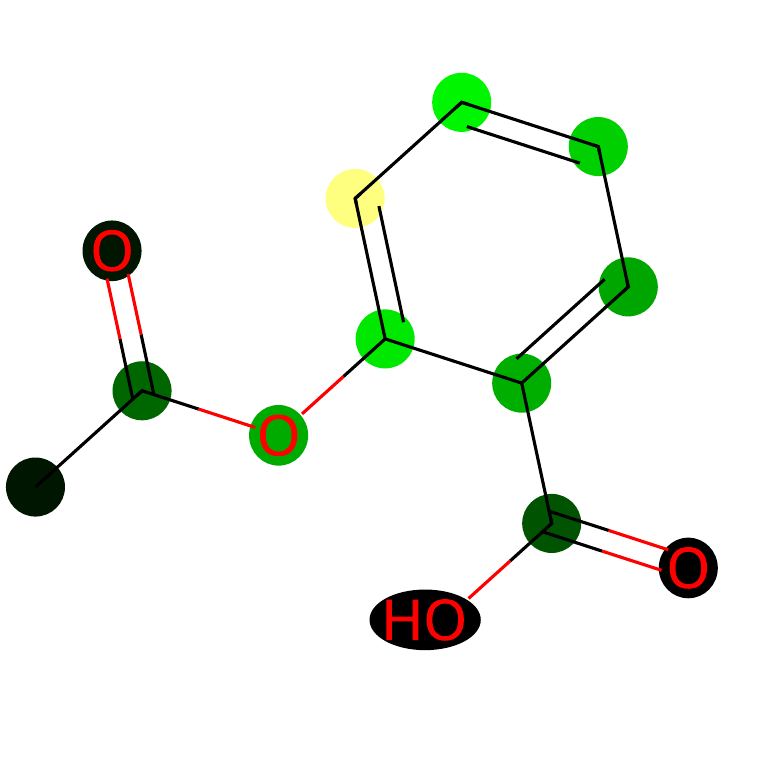}}%
\subfloat[s = 4]{
\label{fig:scale_10}
\includegraphics[scale = 0.2]{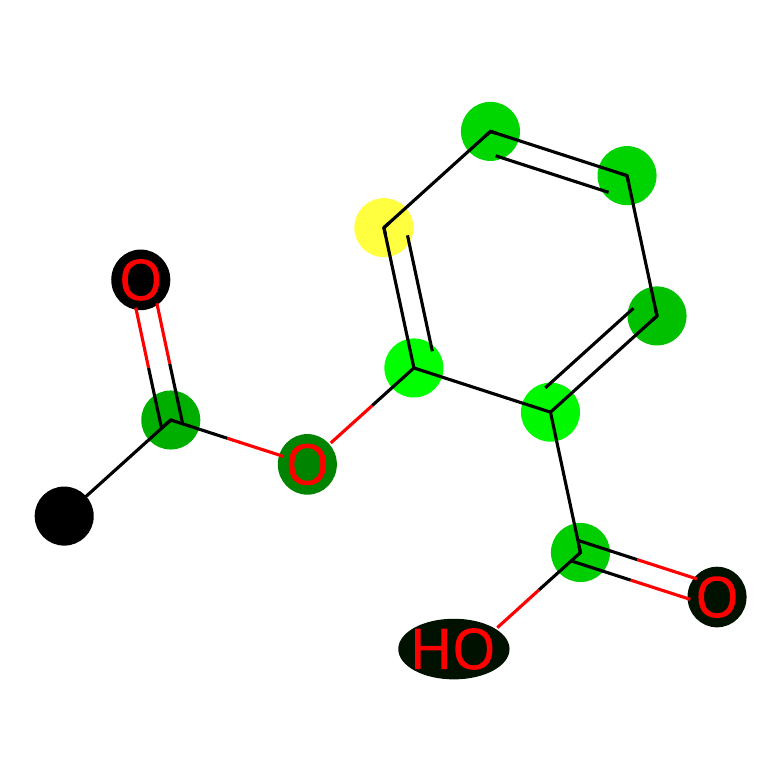}
}
\subfloat{
\label{fig:color_bar}
\includegraphics[scale = 0.2]{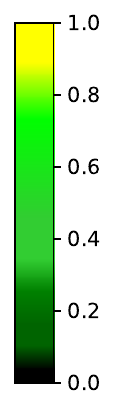}
}
\caption{Visualization of some of the wavelets with scaling parameters on the Aspirin $\text{C}_9\text{H}_8\text{O}_4$ molecular graph with 13 nodes (i.e. heavy atoms). The center node is colored yellow. The colors varying from bright to dark illustrate the diffusion rate from the center node to the others, i.e. nodes that are closer to the center node have brighter colors.
Low-scale wavelets are highly localized, whereas the high-scale wavelets can spread out more nodes on the molecular graphs}
\label{fig:diffusion_wavelet}
\end{figure}

Let $\mathcal{A} \in \mathbb{R} ^ {n\times n}$ be the adjacency matrix of an undirected graph $G = (\mathcal{V}, \mathcal{E})$. The normalized graph Laplacian is defined as $\mathcal{L} = \mathcal{I}_n - \mathcal{D}^{-1/2} \mathcal{A} \mathcal{D} ^ {-1/2}$, where $\mathcal{I}_n$ is the identity matrix and $\mathcal{D}$ is the diagonal matrix of node degrees. $\mathcal{L}$ can be decomposed into a complete set of orthonormal eigenvectors $U = (u_1, u_2, ..., u_n)$ associated with real and non-negative eigenvalues $\{\lambda\}_1^n$. While graph Fourier transform uses $U$ as a set of bases to project the graph signal from the vertex domain to the spectral domain, graph wavelet transform constructs a set of spectral graph wavelets as bases for this projection via: 
\[
\psi_s = U \Sigma_s U^T 
\]
where $\Sigma_s=\text{diag}(g(s\lambda_1), g(s\lambda_2), ..., g(s\lambda_n))$ is a scaling matrix of eigenvalues, $\psi_s=(\psi_{s1}, \psi_{s2}, ..., \psi_{sn})$ and each wavelet $\psi_{si}$ indicates how a signal diffuses away from node $i$ at scale $s$; we choose $g(s\lambda) = e ^ {-s \lambda}$ as a heat kernel \cite{wavelet2018}. Since a node's neighborhoods can be adjusted by varying the scaling parameter $s$ \cite{wavelet}, using multiple sets of wavelets at different scales can provide comprehensive information on the graph's structure. It means that larger values of $s_i$ correspond to larger neighborhoods surrounding a center node. Figure \ref{fig:diffusion_wavelet} illustrates how wavelets can be used to determine neighborhoods at different scales on a molecular graph. In this work, we leverage this property of graph wavelets to generate node positional representations that can capture the structural information of a center node on the graph at different resolutions. We employ $k$ diffusion matrices $\{\psi_{s_i}\}_{i=1}^k$ in which each $\psi_{s_i}$ has a size of $n \times n$, resulting in a tensor of graph 
wavelets $\mathcal{P} \in \mathbb{R} ^ {n \times n \times k}$. Additionally, WavePE is a generalized version of RWPE \cite{rw} as the random walk process can be regarded as a type of discrete diffusion. In the following section, we demonstrate the use of tensor contractions to generate a tensor of node positional representations $\mathbf{P} \in \mathbb{R} ^ {n \times k}$ from $\mathcal{P}$. In general, Fig.\ref{fig:wavepe} demonstrates our wavelet positional encoding method. 

\subsection{Equivariant Encoding}

It is important to note that our spectral graph wavelets computed from the previous section must be further encoded in a permutation-equivariant manner. For simplicity, that means if we permute (i.e. change the order) the set of nodes, 
their position encodings must be transformed accordingly. In this section, we formally define permutation symmetry, i.e. symmetry to the action of the symmetric group, $\mathbb{S}_n$, and construct permutation-equivariant neural networks to encode graph wavelets. An element $\sigma \in \mathbb{S}_n$ is a permutation of order $n$, or a bijective map from $\{1,\ldots, n\}$ to $\{1,\ldots, n\}$. For example, the action of $\mathbb{S}_n$ on an adjacency matrix $\mathcal{A} \in \mathbb{R}^{n \times n}$ and on a latent matrix $\mathcal{Z} \in \mathbb{R}^{n \times d_z}$ are:
\[
[\sigma \cdot \mathcal{A}]_{i_1, i_2} = \mathcal{A}_{\sigma^{-1}(i_1), \sigma^{-1}(i_2)}, \ \ \ \ 
[\sigma \cdot \mathcal{Z}]_{i, j} = \mathcal{Z}_{\sigma^{-1}(i), j},
\]
for $\sigma \in \mathbb{S}_n$. Here, the adjacency matrix $\mathcal{A}$ is a second-order tensor with a single feature channel, while the latent matrix $\mathcal{Z}$ is a first-order tensor with $d_z$ feature channels. In general, the action of $\mathbb{S}_n$ on a $k$-th order tensor $\mathcal{X} \in \mathbb{R}^{n^k \times d}$ (i.e. the last index denotes the feature channels) is defined similarly as:
\[
[\sigma \cdot \mathcal{X}]_{i_1, .., i_k, j} = \mathcal{X}_{\sigma^{-1}(i_1), .., \sigma^{-1}(i_k), j}, \hspace{20pt} \sigma \in \mathbb{S}_n.
\]
\noindent
Formally, we define these equivariant and invariant properties in Def.~\ref{def:Sn-equivariant} and equivariant neural networks in Def.~\ref{def:Sn-network}.

\begin{definition} \label{def:Sn-equivariant}
An $\mathbb{S}_n$-equivariant (or permutation equivariant) function is a function $f\colon \mathbb{R}^{n^k \times d} \rightarrow \mathbb{R}^{n^{k'} \times {d'}}$ that satisfies $f(\sigma \cdot \mathcal{X}) = \sigma \cdot f(\mathcal{X})$ for all $\sigma \in \mathbb{S}_n$ and $\mathcal{X} \in \mathbb{R}^{n^k \times d}$. 
Similarly, we say that $f$ is $\mathbb{S}_n$-invariant (or permutation invariant) if and only if $f(\sigma \cdot \mathcal{X}) = f(\mathcal{X})$.
\end{definition} 

\begin{definition} \label{def:Sn-network}
An $\mathbb{S}_n$-equivariant network is a function $f: \mathbb{R}^{n^k \times d} \rightarrow \mathbb{R}^{n^{k'} \times d'}$ 
defined as a composition of $\mathbb{S}_n$-equivariant linear functions $f_1, .., f_T$ and $\mathbb{S}_n$-equivariant nonlinearities $\gamma_1, .., \gamma_T$:
\[
f = \gamma_T \circ f_T \circ .. \circ \gamma_1 \circ f_1. 
\]
On the another hand, an $\mathbb{S}_n$-invariant network is a function $f: \mathbb{R}^{n^k \times d} \rightarrow \mathbb{R}$ defined as a composition of an $\mathbb{S}_n$-equivariant network $f'$ and an $\mathbb{S}_n$-invariant function on top of it, e.g., $f = f'' \circ f'$.
\end{definition}

\noindent
In order to build permutation-equivariant neural networks, we revisit some basic tensor operations: 
the tensor product $A \otimes B$ (see Def.~\ref{def:product}) and tensor contraction ${A_{\downarrow}}_{x_1, .., x_p}$ (see Def.~\ref{def:contraction}). 
It can be shown that these tensor operations respect permutation equivariance \citep{HyEtAl2018} \citep{Kondor2018}. 

\begin{definition} \label{def:product}
The \textbf{tensor product} of $A \in \mathbb{R}^{n^a}$ with $B \in \mathbb{R}^{n^b}$ yields a tensor $C = A \otimes B \in \mathbb{R}^{n^{a + b}}$ where
\[
C_{i_1, i_2, .., i_{a + b}} = A_{i_1, i_2, .., i_a} B_{i_{a + 1}, i_{a + 2}, .., i_{a + b}}.
\]
\end{definition}

\begin{definition} \label{def:contraction}
The \textbf{contraction} of $A \in \mathbb{R}^{n^a}$ along the pair of dimensions $\{x, y\}$ (assuming $x < y$) yields a $(a - 2)$-th order tensor
\[
C_{i_1, .., i_{x - 1}, j, i_{x + 1}, .., i_{y - 1}, j, i_{y + 1}, .., i_a} = \sum_{i_x, i_y} A_{i_1, .., i_a},
\]
where we assume that $i_x$ and $i_y$ have been removed from amongst the indices of $C$. Using Einstein notation, this can be written more compactly as
\[
C_{\{i_1, i_2, .., i_a\} \setminus \{i_x, i_y\}} = A_{i_1, i_2, .., i_a} \delta^{i_x, i_y},
\]
where $\delta$ is the Kronecker delta. In general, the contraction of $A$ along dimensions $\{x_1, .., x_p\}$ yields a tensor $C = {A_\downarrow}_{x_1, .., x_p} \in \mathbb{R}^{n^{a - p}}$ where
\[
{A_\downarrow}_{x_1, .., x_p} = \sum_{i_{x_1}} \sum_{i_{x_2}} ... \sum_{i_{x_p}} A_{i_1, i_2, .., i_a},
\]
or compactly as
\[
{A_\downarrow}_{x_1, .., x_p} = A_{i_1, i_2, .., i_a} \delta^{i_{x_1}, i_{x_2}, ..,i_{x_p}}.
\]
\end{definition}

Based on these tensor contractions and Def.~\ref{def:Sn-equivariant}, we can construct the second-order $\mathbb{S}_n$-equivariant networks encoding a graph with adjacency matrix $\mathcal{A} \in \mathbb{R}^{n \times n}$, node features $\mathcal{F}_v \in \mathbb{R}^{n \times d_v}$ and edge features $\mathbb{R}^{n \times n \times d_e}$ as in Section ~\ref{exp:2nd-order}: 
$$f = \gamma \circ \mathcal{M}_T \circ .. \circ \gamma \circ \mathcal{M}_1.$$
The ``raw'' graph wavelets can be treated as a second-order tensor of size $n \times n \times k$ where $k$ is the number of scales, similarly as the edge features. We employ the higher-order permutation-equivariant message passing proposed by \cite{maron2018invariant}, \cite{HyEtAl2018} and \cite{Kondor2018} to encode the ``raw'' graph wavelets from size $n \times n \times k$ into $n \times k$ that will be further used as nodes/tokens' embeddings of our Transformer architecture (see Fig.~\ref{fig:overall-idea}).

\subsubsection{Higher-order message passing} \label{exp:2nd-order}

The second order message passing has the message $\mathcal{H}_0 \in \mathbb{R}^{|\mathcal{V}| \times |\mathcal{V}| \times (d_v + d_e)}$ initialized by promoting the node features $\mathcal{F}_v$ to a second order tensor (e.g., we treat node features as self-loop edge features), and concatenating with the edge features $\mathcal{F}_e$. Iteratively,
$$
\mathcal{H}_t = \gamma(\mathcal{M}_t), \ \ \ \ 
\mathcal{M}_t = \mathcal{W}_t \bigg[ \bigoplus_{i, j} (\mathcal{A} \otimes {\mathcal{H}_{t - 1})_\downarrow}_{i, j} \bigg],
$$
where $\mathcal{A} \otimes \mathcal{H}_{t - 1}$ results in a fourth order tensor while $\downarrow_{i, j}$ contracts it down 
to a second order tensor along the $i$-th and $j$-th dimensions, $\oplus$ denotes concatenation along the feature channels, and $\mathcal{W}_t$ denotes a multilayer perceptron on the feature channels. 
We remark that the popular MPNNs \citep{10.5555/3305381.3305512} is a lower-order one and a special case in which $\mathcal{M}_t = \mathcal{D}^{-1}\mathcal{A}\mathcal{H}_{t - 1}\mathcal{W}_{t - 1}$ where $\mathcal{D}_{ii} = \sum_j \mathcal{A}_{ij}$ is the diagonal matrix of node degrees.
The message $\mathcal{H}_T$ of the last iteration is still second order, so we contract it down to the first order latent $\mathcal{Z} = \bigoplus_i {{\mathcal{H}_T}_\downarrow}_i$.

\section{Multiresolution Graph Transformers} \label{sec:method_mgt}
In this section, we present Multiresolution Graph Transformers (MGT), a neural network architecture for learning hierarchical structures. MGT uses Transformers to yield the representations of macromolecules at different resolutions. While previous work either neglects the hierarchical characteristics of large molecules or fails to model global interactions between distant atoms, our proposed approach can satisfy these two properties via multiresolution analysis.

Figs. \ref{fig:atom_level_mgt} and \ref{fig:substructure_level_mgt} show an overview of our framework. MGT consists of three main components: an atom-level encoder, a module to extract substructures, and a substructure-level encoder. We use a graph transformer to generate the atomic embeddings. Then, substructures present in molecules are extracted by a learning-to-cluster algorithm. The molecular graph is coarsened into a set of substructures, and we use a pure Transformer encoder to learn their relations.


\subsection{Atom-Level Encoder}
To utilize the proposed wavelet positional encoding demonstrated in Section \ref{wavelet}, we leverage the design of the graph transformer proposed in \cite{gps}, which is a general, powerful, and scalable Graph Transformer (GraphGPS) for graph representation learning. Let $\mathbf{A} \in \mathbb{R} ^ {n \times n}$ be the adjacency matrix of a graph with $n$ nodes and $e$ edges; $\textbf{X}^l$ and $\textbf{E}^l$ are node and edge features at layer $l$-th, respectively. In addition, $\textbf{X} ^ 0 \in \mathbb{R} ^{n \times d}$ and $\textbf{E}^0 \in \mathbb{R}^ {e \times d}$ are initial atom and bond features embedded in $d$-dimensional spaces created by two embedding layers. The wavelet positional vectors $\textbf{p} \in \mathbb{R} ^ {n \times k}$ are fed to an encoder (e.g., a feed-forward neural network or a linear transformation), yielding a tensor of positional features $\textbf{P} \in \mathbb{R} ^ {n \times d_p}$. We let $\textbf{X}^0:= \text{concat}(\textbf{X}^0, \textbf{P})$ to produce new node features $\textbf{X}^0 \in \mathbb{R} ^ {n \times (d + d_p)}$. From here, we define $d := d + d_p $, and for convenience, the output dimensions of all layers are equal to $d $. 

Each layer of GraphGPS uses a message-passing neural network ($\text{MPNN}^l$) to exchange information (i.e., messages) within the neighborhood and a self-attention layer ($\text{SA}^l$) described in Eq. (\ref{eq:1}) to compute global interactions among distant nodes: 
\begin{align}
    \textbf{X}_L^{l+1}, \textbf{E}^{l+1} & = \text{MPNN} ^ l(\textbf{X}^l, \textbf{E}^l, \textbf{A})\\
    \textbf{X}_G ^ {l+1} &= \text{SA} ^ l(\textbf{X}^{l}) \\
    \textbf{X} ^ {l+1} &= \text{FFN} ^ l(\textbf{X}_L ^ {l+1} + \textbf{X}_G ^ {l+1}) \label{eq:4}
\end{align}
where $\textbf{X}_L ^ {l+1}$ and $\textbf{X}_G ^ {l+1}$ are node local and global representations; they are unified into $\textbf{X} ^ {l+1}$ via Eq. \ref{eq:4}. Popular techniques such as Dropout \cite{dropout} and normalization \cite{batchnorm, layernorm} are omitted for the sake of clarity. By feeding the molecular graph through $L$ layers, we attain two tensors $\textbf{X}_a:=\textbf{X}^L$ and $\textbf{E}_a:=\textbf{E}^L$ indicating the node and edge embeddings, respectively.

\subsection{Learning to Cluster}
In this work, we use a message-passing neural network augmented with differentiable pooling layers \cite{diffpool, Hy_2023} to cluster the atoms into substructures automatically: 
\begin{align}
    \textbf{Z} &= \text{MPNN}_\text{e}(\textbf{X}_a, \textbf{E}_a, \textbf{A}) \label{eq:5} \\
    \textbf{S} &= \text{Softmax}(\text{MPNN}_\text{c}(\textbf{X}_a, \textbf{E}_a, \textbf{A}) )
    \label{eq:6}
\end{align}
where $\text{MPNN}_\text{e}$ and $\text{MPNN}_\text{c}$ are two-layer message-passing networks that learn to generate node embeddings ($\textbf{Z} \in \mathbb{R} ^ {n \times d}$) and a clustering matrix ($\textbf{S} \in \mathbb{R} ^ {n \times C})$, respectively; $C$ denotes the number of substructures in molecules. A tensor of features $\textbf{X}_s \in \mathbb{R} ^ {C \times d}$ for the substructures is computed: 
\begin{equation}
    \textbf{X}_s = \textbf{S} ^ T \textbf{Z}
\end{equation}
This learning-to-cluster module is placed after the atom-level encoder. Intuitively, atom nodes updated with both local and global information should be classified into accurate substructures.

\subsection{Substructure-level Encoder}
Given a set of substructures $\mathcal{V}_s$ with a tensor of features $\mathbf{X}_s \in \mathbb{R} ^ {C \times d}$, we forward $\mathbf{X}_s$ to $L$ conventional Transformer encoder layers \cite{vaswari} to capture their pairwise semantics:
\begin{align}
    \textbf{H}_\text{1}^{l+1} &= \text{Norm}(\text{SA}^l(\textbf{H} ^ l) + \textbf{H} ^ l)\\
    \textbf{H} ^ {l+1} &= \text{Norm}(\text{FFN}(\textbf{H}_\text{1}^{l+1}) + \textbf{H}_\text{1}^{l+1}) 
\end{align}
where $\text{SA}$ refers to (multi-head) self-attention described in Eq. (\ref{eq:1}), and $\textbf{H}^0$ is equal to $\textbf{X}_s$. Additionally, we add a long-range skip connection to alleviate gradient vanishing as:
\begin{equation}
    \textbf{H}_s = \text{FFN}(\text{concat}(\textbf{H}^0, \textbf{H}^{L}))
\end{equation}
$\textbf{H}_s \in \mathbb{R} ^ {C \times d}$ is the output indicating the representations for the substructures. Finally, we aggregate all $C$ vectors $h_s \in \textbf{H}_s$ to result in a unique representation $z \in \mathbb{R} ^ d$ for the molecules (refer to Section \ref{heirachical_coarsen}), before feeding it to a feed-forward network to compute the final output $y \in \mathbb{R} ^ c$ for property prediction:
\begin{align}
    z &= \zeta (\{h_s\}_{s = 1} ^ C) \\
    \hat{y} &= \text{FFN}(z) 
\end{align}

\paragraph{Training Objective} We train MGT by minimizing $\mathcal{L}$:
\begin{equation}
    \mathcal{L} = \mathcal{L}_1 + \lambda_1 \mathcal{L}_{LP} + \lambda_2 \mathcal{L}_{E} 
\end{equation}
where $\mathcal{L}_1 = l(\hat{y}, y)$ denotes the loss function between predicted values and ground truths (e.g., cross-entropy or mean-squared error), $\mathcal{L}_{LP} =||\textbf{A} - \textbf{S} \textbf{S} ^ T||_F $ indicates auxiliary link prediction loss ($||\cdot||_F$ denotes the Frobenius norm), and $L_E = \frac{1}{n} \sum_{i=1}^n H(\textbf{S}_i)$ denotes the entropy regularization of the cluster assignment, i.e. each atom should be assigned into a unique cluster. Additionally, $\lambda_1$ and $\lambda_2$ are hyperarameters. 
%

%% file: experiments.tex
\begin{table*}[t]
    \centering
    \caption{Experimental results on the polymer property prediction task. All the methods are trained in four different random seeds and evaluated by MAE $\downarrow$. Our methods are able to attain better performance across three DFT properties of polymers while having less number of parameters. All the properties are measured in eV.}
    \vskip 0.1in
    \begin{tabular}{ l c c c c  } 
        \toprule
        \multirow{2}{*}{Model} & \multirow{2}{*}{No. Params}&\multicolumn{3}{c}{Property}  \\
        \cmidrule(lr){3-5}
        & & GAP  & HOMO & LUMO \\
        \midrule
        DFT error & & 1.2 & 2.0 & 2.6 \\
        Chemical accuracy & & 0.043 & 0.043 & 0.043 \\
        \midrule
        GCN & 527k & 0.1094 $\pm$ 0.0020 & 	0.0648 $\pm$ 0.0005 & 0.0864 $\pm$ 0.0014 \\
        GCN + Virtual Node & 557k & 0.0589 $\pm$ 0.0004 & 0.0458 $\pm$ 0.0007 & 0.0482 $\pm$ 0.0010  \\
        GINE & 527k & 0.1018 $\pm$ 0.0026 & 0.0749 $\pm$ 0.0042 & 0.0764 $\pm$ 0.0028 \\
        GINE + Virtual Node & 557k & 0.0870 $\pm$ 0.0040 & 0.0565 $\pm$ 0.0050 & 0.0524 $\pm$ 0.0010  \\
        \midrule
        GPS  & 600k & 0.0467 $\pm$ 0.0010 & 0.0322 $\pm$ 0.0020 & 0.0385 $\pm$ 0.0006 \\
        Transformer + LapPE & 700k & 0.2949 $\pm$ 0.0481 & 0.1200 $\pm$ 0.0206 & 0.1547 $\pm$ 0.0127\\
\midrule
        MGT + LapPE (ours) & 499k & \textbf{0.0378 $\pm$ 0.0004} & \textbf{0.0270 $\pm$ 0.0010} & 0.030 $\pm$ 0.0006 \\
        MGT + RWPE (ours) & 499k & 0.0384 $\pm$ 0.0015 & 0.0274 $\pm$ 0.0005 & \textbf{0.0290 $\pm$ 0.0007} \\
        MGT + WavePE (ours) & 499k & 0.0387 $\pm$ 0.0011 & 0.0283 $\pm$ 0.0004 & 0.0290 $\pm$ 0.0010 \\
        \bottomrule
    \end{tabular}
    \label{tab:polymer_result}

\end{table*}

\section{Experiments} \label{sec:experiments}
\begin{figure}%
\centering
\captionsetup[subfloat]{captionskip=1mm} 
\subfloat[]{%
\label{fig:peptides}%
\includegraphics[scale = 0.55]{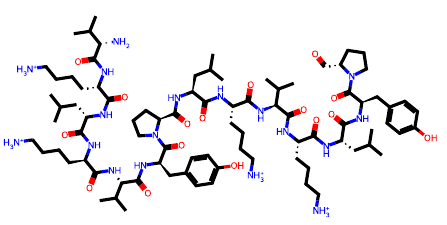}}%
\qquad
\subfloat[]{%
\label{fig:polymer}%
\includegraphics[scale = 0.4]{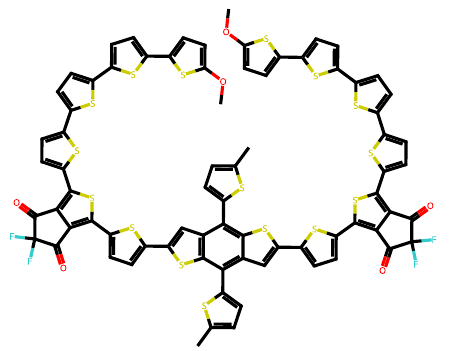}}%
\caption{Examples of two macromolecules. a) An example of a peptide that consists of many functional groups. b) An example of a polymer that consists of many repeating units}
\label{fig:macromolecules}
\end{figure}
We empirically validate our proposed approach in two types of macromolecules including peptides and polymers. Figure \ref{fig:macromolecules} illustrates two examples of macromolecules in the datasets. Our PyTorch implementation is publicly available at \url{https://github.com/HySonLab/Multires-Graph-Transformer}.

\subsection{Polymer Property Prediction}
In modern quantum chemistry, density functional theory (DFT) is the most widely used and successful approach for computing the electronic structure of matter. Even though DFT is able to calculate many properties of molecular systems with high accuracy, the computational cost is significantly expensive, especially for macromolecules with hundreds or thousands of atoms. Polymers are long chains of repetitive substructures known as repeating units. They, as a result, are also hierarchically structured and contain various types of long-range dependencies among the atoms. Since polymers have a wide range of applications in daily life, it is essential to understand their molecular properties. In this section, we demonstrate the efficacy of our proposed model, MGT, along with atomic positional encoding in estimating DFT calculation on polymers.

\paragraph{Experimental Setup} We use a polymer dataset proposed in \cite{st2019message}. Each polymer is associated with three types of density functional theory (DFT) \cite{PhysRev.136.B864} properties including the first excitation energy of the monomer calculated with time-dependent DFT (GAP), the energy of the highest occupied molecular orbital for the monomer (HOMO), and the lowest unoccupied molecular orbital of the monomer (LUMO). 
The dataset is split into train/validation/test subsets with a ratio of 8:1:1, respectively. For training, we normalize all learning targets with a mean of 0 and a standard deviation of 1. 
\paragraph{Baselines and Implementation Details}
As there are no existing baselines on this dataset, we perform experiments with four different models for comparisons. For local GNNs, we use GCN \cite{gcn} and GINE \cite{gin, strategies_pretrain} augmented with virtual nodes as the baselines \footnote{The implementation of local GNN models is taken from https://github.com/snap-stanford/ogb/tree/master/ogb}. Moreover, we use standard Transformer \cite{vaswari} with Laplacian positional encoding \cite{graph_generalize} and GPS \cite{gps} as the baselines for Transformer-based architectures. The implementation of MGT is similar to the Peptide tasks. 
 Please refer to Appendix \ref{ref:appendidx_polymer} for the baseline details. For fair comparisons, all the models are trained in 50 epochs with a learning rate of 0.001 and batch size of 128

 \paragraph{Results}
As shown in Table \ref{tab:polymer_result}, our MGT models achieve the lowest MAE scores across three properties. In addition, WavePE can attain comparable results with LapPE and RWPE for this task. We observe that the vanilla Transformer has the poorest performance. This demonstrates that computing global information without the awareness of locality is not sufficient for macromolecular modeling. As described in Section \ref{sec:method_mgt}, MGT is an extended version of GPS. In particular, a learning-to-cluster module and a substructure-level Transformer encoder are extensions to GPS. The better performance of MGT, as a result, indicates that our methodology in modeling hierarchical structures is appropriate and reasonable.

There are two important benchmark error levels: (1) ``DFT error'', the estimated average error of the DFT approximation to nature; and (2) ``chemical accuracy'', the target error that has been established by the chemistry community. Estimates of DFT error and chemical accuracy are provided by \cite{faber2017machine}. Our model is the only one to achieve the chemical accuracy for all the three molecular properties.

\begin{table*}[t]
    \centering
    \caption{Results on peptides property prediction}
    \vskip 0.15in
    \begin{tabular}{ l c c c } 
        \toprule
        \multirow{2}{*}{Model} & \multirow{2}{*}{No.Params} &\multicolumn{1}{c}{Peptides-struct} & \multicolumn{1}{c}{Peptides-func} \\
        \cmidrule(lr){3-3} \cmidrule(lr){4-4} 
       & & MAE $\downarrow$ & AP $\uparrow$ \\
        \midrule
        GCN & 508k & 0.3496 $\pm$ 0.0013 & 0.5930 $\pm$ 0.0023\\
        GINE & 476k & 0.3547 $\pm$ 0.0045 & 0.5498 $\pm$ 0.0079\\
        GatedGCN & 509k & 0.3420 $\pm$ 0.0013 & 0.5864 $\pm$ 0.0077\\
        GatedGCN + RWPE & 506k & 0.3357 $\pm$ 0.0006 & 0.6069 $\pm$ 0.0035 \\
        \midrule
        Transformer + LapPE & 488k & 0.2529 $\pm$ 0.0016 & 0.6326 $\pm$ 0.0126 \\
        SAN + LapPE & 493k & 0.2683 $\pm$ 0.0043 & 0.6384 $\pm$ 0.0121 \\
        SAN + RWPE & 500k & 0.2545 $\pm$ 0.0012 & 0.6562 $\pm$ 0.0075\\
    \midrule
    MGT + LapPE (ours) & 499k & 0.2488 $\pm$ 0.0014 & 0.6728 $\pm$ 0.0152 \\
    MGT + RWPE (ours) & 499k  & 0.2496 $\pm$ 0.0009 & 0.6709 $\pm$ 0.0083\\ 
    MGT + WavePE (ours) & 499k & \textbf{0.2453 $\pm$ 0.0025} & \textbf{0.6817 $\pm$ 0.0064} \\
        \bottomrule
    \end{tabular}

    \label{tab:peptides_results}
\end{table*}

\subsection{Peptides Property Prediction}

Peptides are small chains of amino acids found in nature that serve a variety of critical biological roles \cite{peptides}; however, they are far shorter than proteins. Because each amino acid is made up of several heavy atoms, a peptide's molecular graph is significantly greater than that of a small drug-like molecule. Since peptides are formed by sequences of amino acids, they are naturally hierarchical and long-range dependencies \cite{long_range}, i.e. a peptide should be ideally segmented into an exact set of amino acids. Therefore, we evaluate our method on peptide structures to demonstrate its superiority. 

\paragraph{Experimental Setup}
We run experiments on two real-world datasets including (1) Peptides-struct and (2) Peptides-func \cite{long_range}. The two datasets are multi-label graph classification problems and share the same peptide molecular graphs, but with different tasks. While the former consists of 10 classes based on peptides function, the latter is used to predict 11 aggregated 3D properties of peptides at the graph level. 
For a fair comparison, we follow the experimental and evaluation setting of \cite{long_range} with the same train/test split ratio. We use mean absolute error (MAE) and average precision (AP) to evaluate the method's performance for Peptides-struct and Peptides-func, respectively. 
\paragraph{Baselines and Implementation Details}
We compare our proposed approach with the baselines from \cite{long_range}. The local message-passing network class involves GCN \cite{gcn}, GCNII \cite{gated_gcn}, GINE \cite{gin, strategies_pretrain}, and GatedGCN \cite{gated_gcn}. For Transformer-based architectures, we compare our method with vanilla Transformer \cite{vaswari} with Laplacian PE \cite{benchmark_graph, graph_generalize} and SAN \cite{san}. Since all baselines are limited to approximately 500k learnable parameters, we also restrict MGT to roughly the same number of parameters. Additionally, we use GatedGCN \cite{gated_gcn} for local message passing customized with the PEG technique to stabilize the positional features \cite{wang2022equivariant} \footnote{The implementation of GPS is adapted from https://github.com/vijaydwivedi75/lrgb.git.}. We experiment with each task in four different random seeds. We provide further implementation details of MGT for this task in Appendix \ref{ref:appendix_peptides}. 

\paragraph{Results}
Table \ref{tab:peptides_results} shows that our proposed MGT + WavePE achieves the best performances in two peptide prediction tasks. In addition to WavePE, MGT + RWPE also attains the second-best performances. The superiority of WavePE to RWPE can be explained as mentioned in Section \ref{wavelet} that WavePE is a generalized version of RWPE. In particular, our proposed MGT outperforms all the baselines in the Petides-func task by a large margin and decreases the MAE score to less than 0.25 in the Petides-struct task.

\subsection{Drug-like molecule property prediction}

\begin{table}[h]
    \centering
    \caption{Experimental results on the ZINC-12K dataset}
    \begin{tabular}{ l  c c   } 
        \toprule
        Method & No. Params & MAE $\downarrow$ \\
        \midrule 
        GCN &  505k & 0.367 $\pm$ 0.011 \\
        GINE & 510k & 0.526 $\pm$ 0.051 \\
        GAT &  531k & 0.384 $\pm$ 0.007 \\ 
        PNA &  387k & 0.142 $\pm$ 0.010 \\
        MPNN & 418k & 0.145 $\pm$ 0.007 \\
        GatedGCN & 505k & 0.214 $\pm$ 0.006 \\
        SAN & 509k & 0.139 $\pm$ 0.006 \\
        Graphormer & 489k & 0.122 $\pm$ 0.006 \\
        GPS &   - & 0.070 $\pm$ 0.004  \\ 
        Spec-GN & 503k & 0.0698 $\pm$ 0.002 \\
        \midrule
         MGT + WavePE (ours) & 499k & 0.131 $\pm$ 0.003 \\
        \bottomrule
    \end{tabular}
    \label{tab:zinc_experimental_results}
\end{table}
 
Although MGT is intentionally designed for learning to represent hierarchical structures, we report its experimental results on the ZINC-12K dataset \cite{Sterling2015}, which consists of small drug-like molecules, in this section. We train MGT to predict the solubility (LogP) of the molecules with up to 50 heavy atoms on a subset of the ZINC dataset. We follow the split of 10K/1K/1K for training/validation/testing proposed in \cite{benchmark_graph}. Baseline results include GCN \cite{gcn}, GINE \cite{gin, benchmark_graph}, , GAT \cite{gat}, Spec-GN \cite{pmlr-v162-yang22n}, PNA \cite{pna}, GatedGCN \cite{gated_gcn}, GPS \cite{gps}, MPNN \cite{10.5555/3305381.3305512}, SAN \cite{san}, DGN \cite{deeper_gcn}, and Graphormer \cite{graphormer}. According to Table \ref{tab:zinc_experimental_results}, MGT + WavePE outperforms 7 out of 10 other baselines.

\subsection{Visualization}
We use the t-SNE algorithm \cite{tsne} to project the representations produced by MGT (with WavePE) of peptides and polymers of the test datasets into two-dimensional spaces for visualization. Also, we take the probabilistic clustering matrix $\mathbf{S}$ in Eq. (\ref{eq:6}) to visualize the clustering results on the molecules. Specifically, we use the RDKit package \footnote{RDKit: Open-source cheminformatics. https://www.rdkit.org} to draw the molecules.

\begin{figure*}
    \centering
    \includegraphics[scale=0.35]{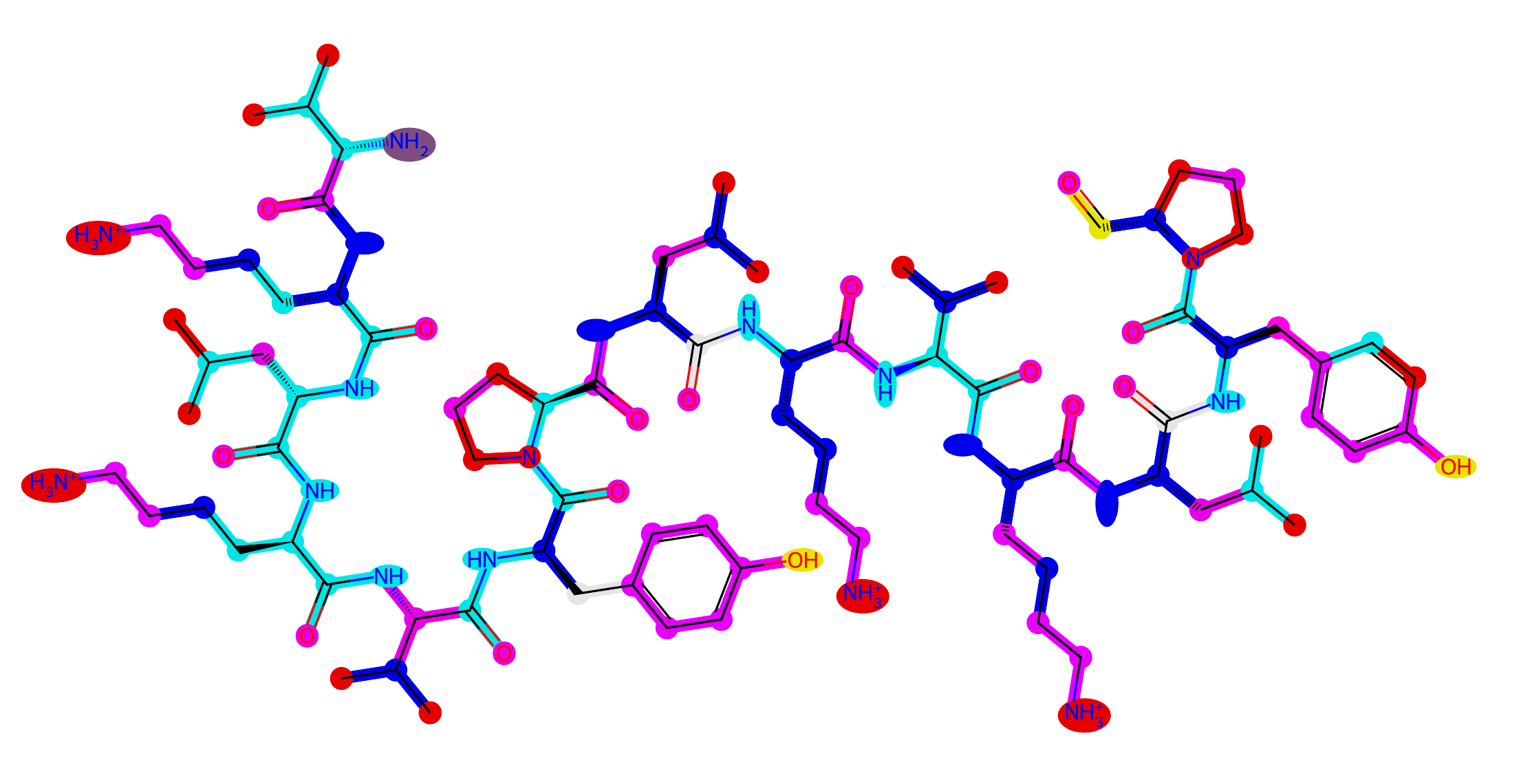}
    \caption{The clustering result on a peptide. MGT can group the atoms of a long peptide into different substructure types. Specifically, the groups NH3 and OH are recognized even though the atoms are located distantly. Also, local rings or segments are also detected.}
    \label{fig:peptides_clustering}
\end{figure*}

\begin{figure*}
    \centering
    \includegraphics[scale = 0.3]{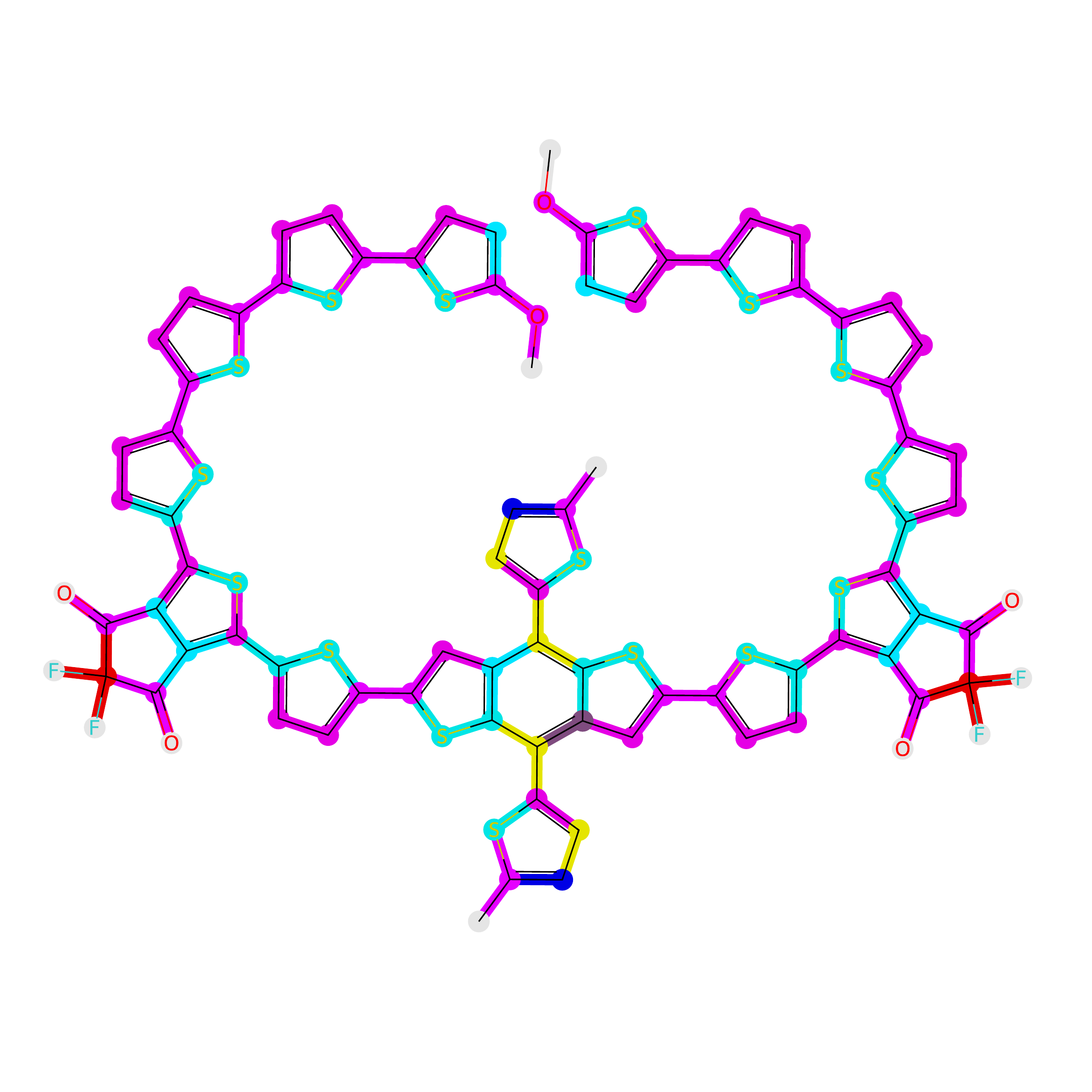}
    \vspace{-20mm}
    \caption{The clustering result on a polymer. By learning to cluster and using a
substructure-level Transformer encoder, MGT can model repetitive patterns existing in polymers. In this example, the model can recognize repeating units in a long-chain polymer or even symmetries.}
    \label{fig:polymer_clustering}
\end{figure*}

\begin{figure*}%
\centering
\subfloat[]{%
\includegraphics[scale = 0.4]{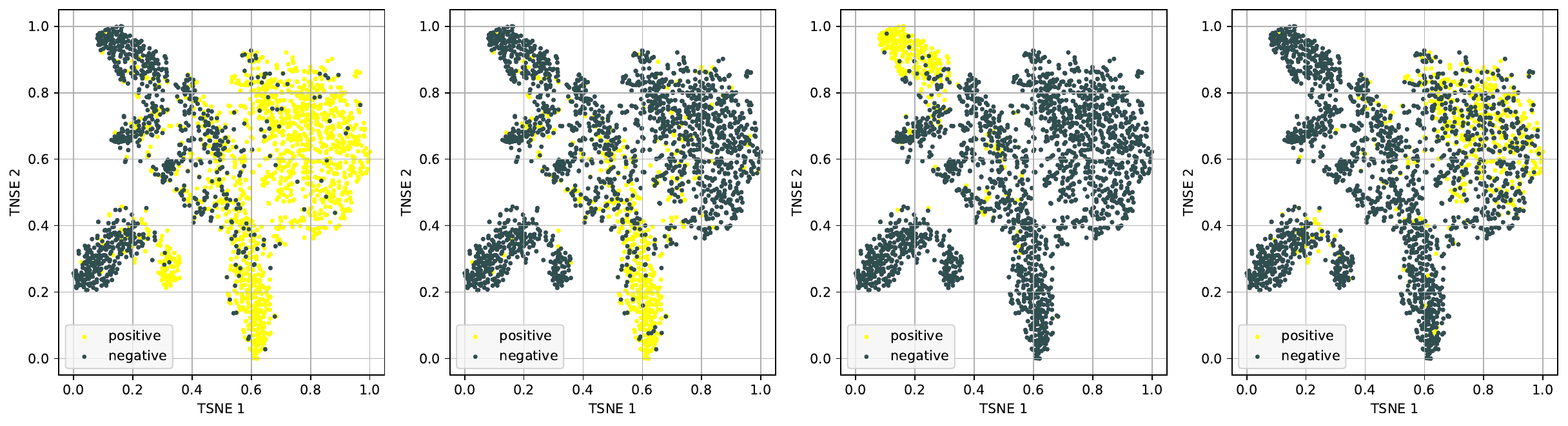}}%
\qquad
\subfloat[]{%
\includegraphics[scale = 0.38]{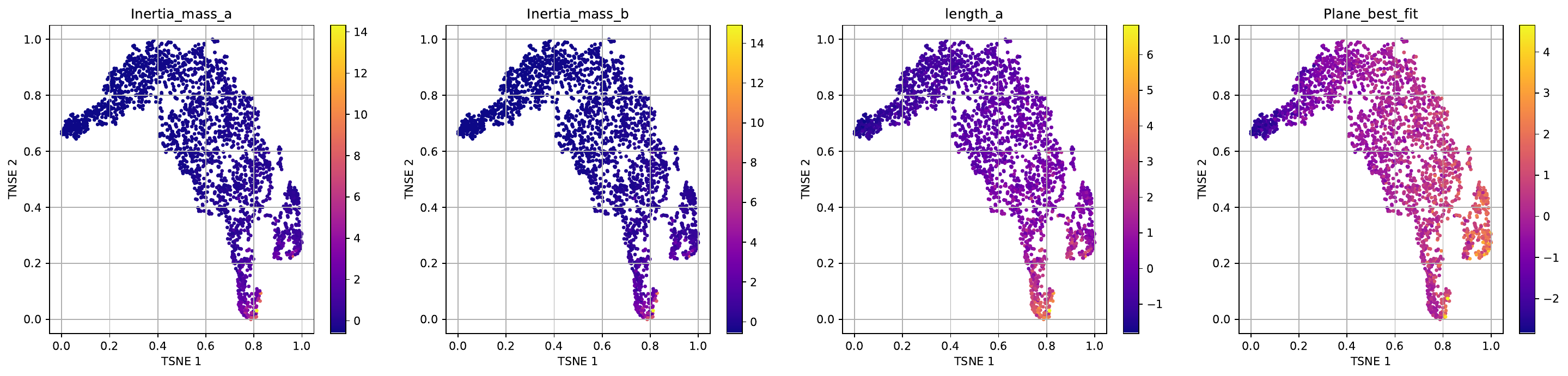}}%
\caption{Visualization of low-dimensional spaces of peptides on two property prediction tasks: Peptides-func and Peptides-struct. All the vectors are normalized to range $[0,1]$. a) t-SNE projection of peptides taken from the Peptides-func testing dataset. We take four random peptide functions, and each figure corresponds to one of the properties with positive (1) and negative (0) ground truths. b) Similarly, we plot the figures of four random peptide properties taken from the Peptides-struct testing dataset. The spectrums represent continuous ground truths, where lower values correspond to cooler colors.}
\label{fig:peptides_rep}
\end{figure*}

\begin{figure*}
    \centering
    \includegraphics[scale=0.37]{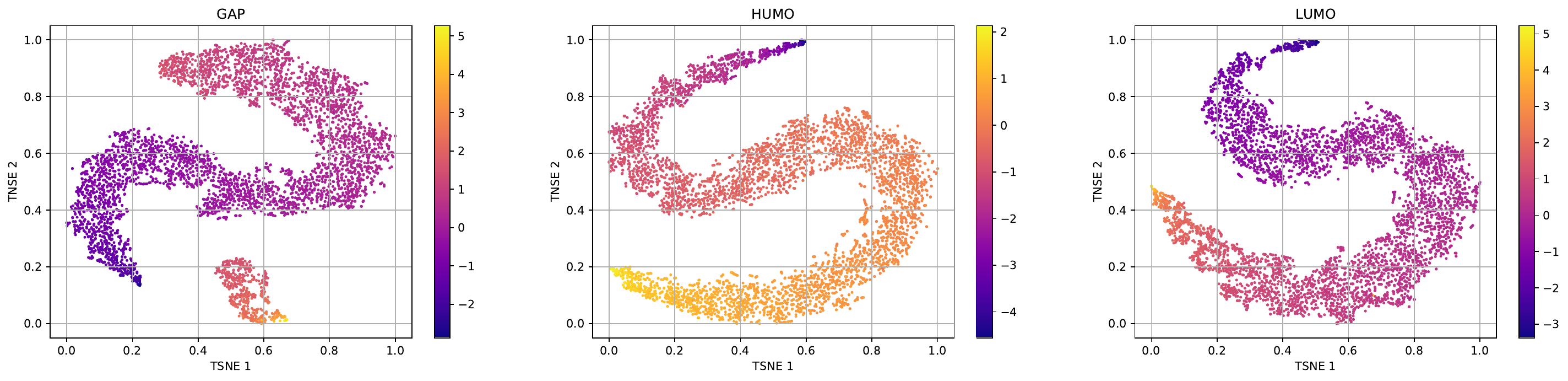}
    \caption{
        t-SNE projection of representations of the test polymers in the dataset. We plot the figures of three properties, including GAP, HUMO, and LUMO. As the labels are continuous values, points are color-coded by spectrums wherein higher values correspond to warmer palettes.
    }
    \label{fig:polymer_rep}
\end{figure*}

Figs. \ref{fig:peptides_rep} and \ref{fig:polymer_rep} show clear and smooth clustering patterns in low-dimensional spaces that indicate our proposed approaches are able to learn meaningful molecular representations for hierarchical structures such as peptides and polymers. Furthermore, according to Figs. \ref{fig:peptides_clustering} and \ref{fig:polymer_clustering}, our learning-to-cluster algorithm and multiresolution analysis can pick up functional groups (for proteins/peptides) and repeating units (for polymers) via back-propagation.



%% file: conclusion.tex
\section{Conclusion} \label{sec:conclusion}
In this paper, we introduce a novel architecture, Multiresolution Graph Transformer (MGT), that is able to learn and capture the molecular structure of macromolecules at multiple levels of resolution. We utilize the popular Transformer architecture to model the long-range atomic interaction. Our proposed model employs a learning-to-cluster algorithm that is trainable via back-propagation in order to construct the hierarchy of coarse-graining graphs (i.e. multiresolution) while detecting important functional groups. Furthermore, to empower MGT, we propose a new atomic positional encoding named WavePE based on multiresolution analysis and wavelet theory. We have shown competitive experimental results on two macromolecule datasets of polymers and peptides, and one small molecule dataset of drug-like compounds. Noticeably, our model achieves chemical accuracy in approximating the density functional theory (DFT) calculation and outperforms other state-of-the-art graph learning methods in the polymers dataset. We have released our software and data publicly. We believe our model and implementation will certainly advance the field of DFT approximation for large-scale molecular structures and allow several downstream applications in drug discovery and materials science.

%% file: acknowledgements.tex
\section*{Acknowledgements}

We would like to thank Dao Quang Huy at FPT Software AI Center, Le My Linh at University of California, Santa Barbara, Andrew Hands at University of Chicago, and Le Duy Dung (Andrew) at VinUniversity for our scientific discussions and their valuable suggestions.

%% file: appendix.tex
\appendix
\section{Implementation Details}
\subsection{Multiresolution Graph Transformer}
\label{ref:appendix_peptides}
In this section, we elaborate on the architecture and hyperparameters used to train and evaluate our MGT to achieve the above numerical results. Table \ref{tab:hyperparameters} show details of the hyperparameters used for MGT in all the experiments. In particular, we use the atom and bond encoder modules provided by OGB \cite{hu2021ogblsc} to attribute the molecular graph. We use two GPS layers to compute the atom-level representations and two Transformer layers for calculating the substructure-level representations. For learning to cluster, we use a 2-layer message-passing network to compute $\textbf{Z}$ and $\textbf{S}$ mentioned in Eq. (\ref{eq:5}) (\ref{eq:6}) as follows:
\begin{align}
    \textbf{Z}_a ^ 1, \textbf{E}_a ^1 &= \text{GatedGCN}^1(\textbf{X}_a, \textbf{E}_a, \textbf{A}) \\
    \textbf{Z}_a ^1 &= \text{Batchnorm}(\text{ReLU}(\textbf{Z}_a ^ 1)) \\
    \textbf{Z}_a ^ 2, \textbf{E}_a ^2 &= \text{GatedGCN}^2(\textbf{Z}_a ^ 1, \textbf{E}_a ^ 1, \textbf{A}) \\
    \textbf{Z}_a ^2 &= \text{Batchnorm}(\text{ReLU}(\textbf{Z}_a ^ 2)) \\
    \textbf{Z} &= \text{concat}(\textbf{Z}_a ^ 1, \textbf{Z}_a ^ 2) \\
    \textbf{Z} &= \text{FFN}(\textbf{Z})
\end{align}
$\textbf{S}$ is computed similarly with an auxiliary Softmax operation on the output to produce a probabilistic clustering matrix.

\begin{table}[h]
    \centering
    \caption{The hyperparameters for MGT}
    \begin{tabular}{ l c   } 
        \toprule
        Hyperparameters & Values \\
        \midrule 
        No. Epoch & 200 \\
        Emb Dim & 84 \\
        Batch size & 128 \\
        Learning rate & 0.001 \\
        Dropout & 0.25 \\
        Attention Dropout & 0.5 \\
        Diffusion Step (K) & [1, 2, 3, 4, 5] \\
        No. Head & 4 \\
        Activation & ReLU \\
        Normalization & Batchnorm \\
        No. Cluster & 10 \\
        $\lambda_1$ & 0.001 \\
        $\lambda_2$ & 0.001 \\
        \bottomrule
    \end{tabular}
    \label{tab:hyperparameters}
\end{table}

\subsection{Baselines used in Polymer Property Prediction}
\label{ref:appendidx_polymer}
Table \ref{tab:baselines_settings} shows the implementation of the baselines we used in the polymer experiments. All the models are designed to have approximately 500 to 700k learnable parameters. For fair comparisons, all the models are trained in 50 epochs with a learning rate of 0.001 and batch size of 128.

\begin{table}[h]
    \centering
    \caption{The detailed settings of baselines for polymer property prediction}
    \begin{tabular}{ l c c c   } 
        \toprule
        Model & No. Layer & Embed Dim & No. Params\\
        \midrule 
        GCN & 5 & 300 & 527k \\
        GCN + Virtual Node & 5 & 156 & 557k \\
        GINE & 5 & 156 & 527k \\
        GINE + Virtual Node & 5 & 120 & 557k \\
        GPS & 3 & 120 & 600k \\
        Transformer + LapPE & 6 & 120 & 700k \\
        \bottomrule
    \end{tabular}
    \label{tab:baselines_settings}
\end{table}